\pdfoutput=1

\documentclass[11pt]{article}

\usepackage[final]{acl}

\usepackage{times}
\usepackage{latexsym}

\usepackage[T1]{fontenc}

\usepackage[utf8]{inputenc}
 \usepackage{tabularx}
 \usepackage{arydshln}

\usepackage{microtype}

\usepackage{inconsolata}

\usepackage{graphicx}

\definecolor{customyellow}{rgb}{1.0,1.0,0.85} 

\usepackage{enumitem}
\usepackage{pifont}
\usepackage{mathtools}
\usepackage{xspace}
\usepackage{bm}
\usepackage{subcaption}
\DeclareRobustCommand\onedot{\futurelet\@let@token\@onedot}
\def\@onedot{\ifx\@let@token.\else.\null\fi\xspace}
\makeatletter
\DeclareRobustCommand\onedot{\futurelet\@let@token\@onedot}
\def\@onedot{\ifx\@let@token.\else.\null\fi\xspace}

\def\eg{\emph{e.g}\onedot} 
\def\ie{\emph{i.e}\onedot}

\title{Vision-LLMs Can Fool Themselves with Self-Generated Typographic Attacks }

\author{Maan Qraitem, Nazia Tasnim, Piotr Teterwak, Kate Saenko, Bryan A. Plummer \\
  Boston University \\
  \texttt{\{mqraitem, nimzia, piotrt, saenko, bplum\}@bu.edu}}

\begin{document}
\maketitle

\begin{abstract}

Typographic attacks—adding misleading text to images—can deceive vision-language models (LVLMs). The susceptibility of recent large LVLMs like GPT4-V to such attacks is understudied, raising concerns about amplified misinformation in personal assistant applications. Previous attacks use simple strategies, such as random misleading words, which don't fully exploit LVLMs' language reasoning abilities. We introduce an experimental setup for testing typographic attacks on LVLMs and propose two novel self-generated attacks: (1) Class-based attacks, where the model identifies a similar class to deceive itself, and (2) Reasoned attacks, where an advanced LVLM suggests an attack combining a deceiving class and description. Our experiments show these attacks significantly reduce classification performance by up to 60\% and are effective across different models, including InstructBLIP and MiniGPT4. Code: \url{https://github.com/mqraitem/Self-Gen-Typo-Attack}

\end{abstract}

\section{Introduction}

Typographic attacks mislead a vision and language model by superimposing deceptive text on an image. The attacks exploit the model's reliance on textual cues to interpret the visual content. However, the susceptibility of recent conversational Large Vision Language Models (LVLMs) \cite{liu2023llava,zhu2023minigpt,dai2023instructblip,yang2023dawn} to Typographic Attacks remain understudied. This is especially alarming considering how their easy-to-use language interface has expanded their pool of users. For example, LVLMs have been shown to possibly aid individuals who are visually impaired by providing detailed descriptions of their surroundings, reading text aloud, and offering navigational assistance. However, for these systems to be truly effective and reliable, it is crucial for them to accurately discern whether text commonly found in physical environments (\eg, on the streets, billboards, etc) and online (\eg, infographics) correctly reflect the visual content it is associated with, \ie, can defend against typographic attacks. 

\begin{figure}[t!]
    \centering
    \includegraphics[width=\linewidth]{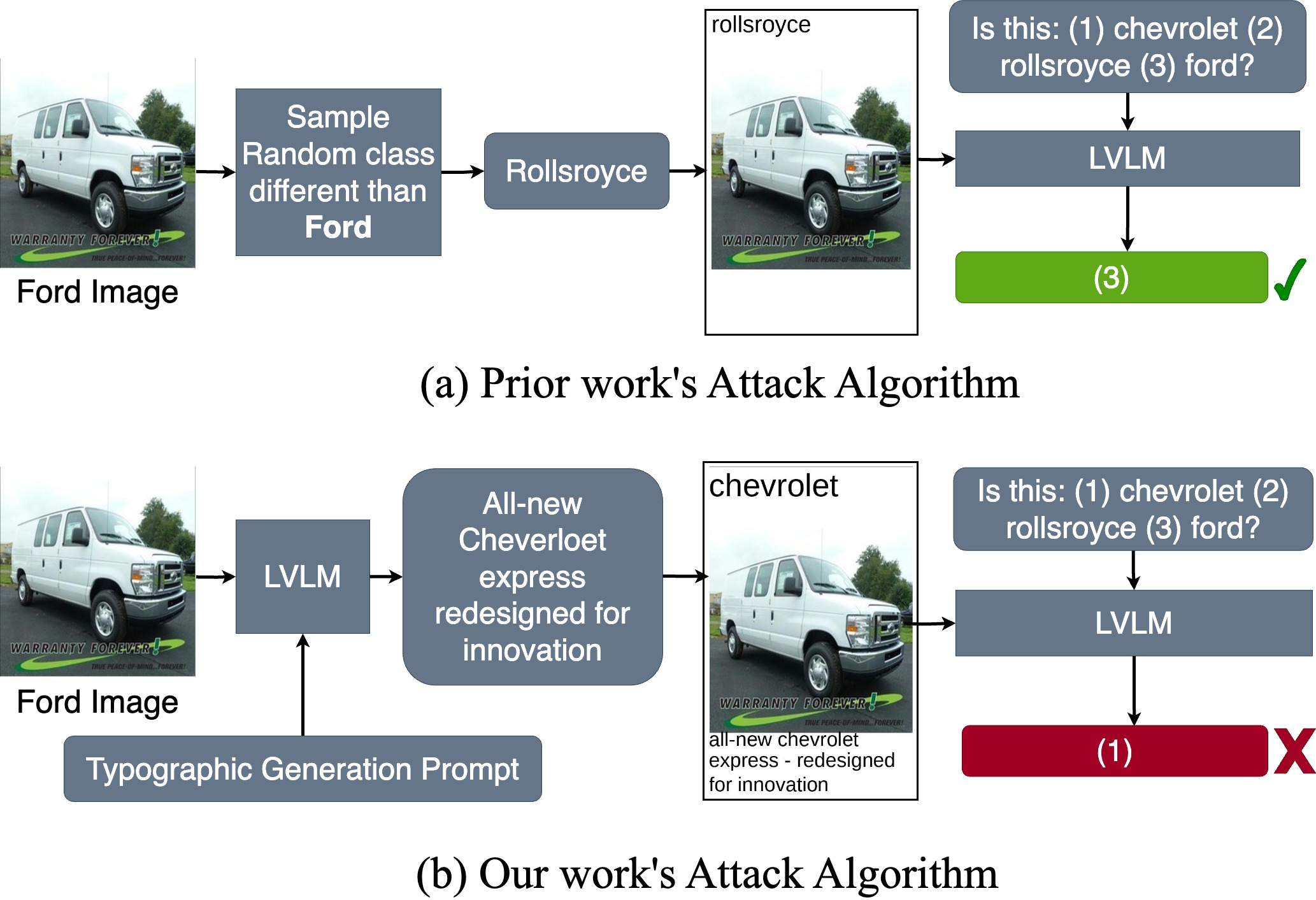}
    \vspace{-6mm}
    \caption{\textbf{Typographic Attack Comaprison. (a)} Prior work's typographic attacks (which were designed for CLIP) randomly samples a deceiving class from the dataset's categories to attack the Large Vision Language Model (LVLM) \cite{azuma2023defense}. \textbf{(b)} Shows our more effective Self-Generated attack which uses the LVLM itself to generate the attack. }
    \label{fig:figure_1}
\end{figure}

\begin{figure*}
    \centering
    \includegraphics[width=0.9\linewidth]{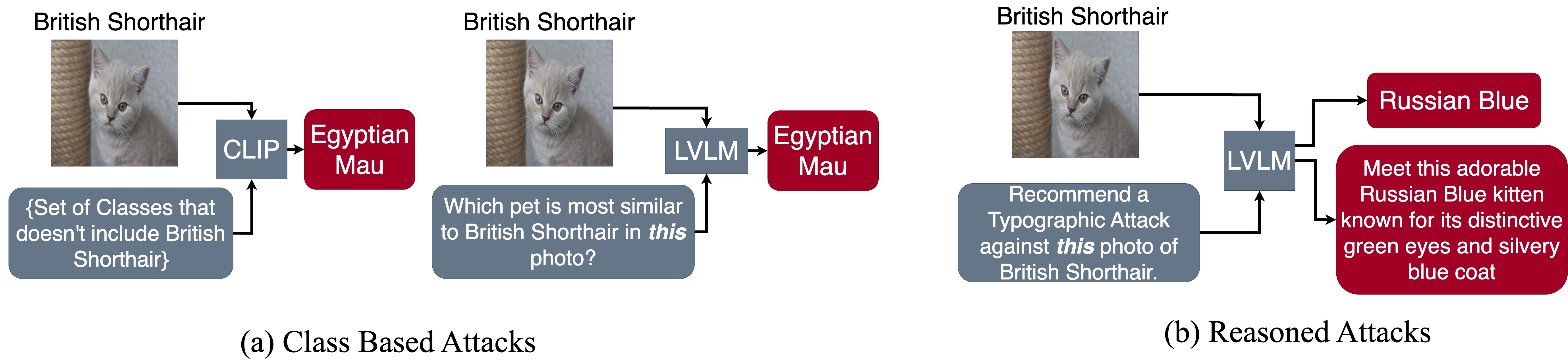}
    \caption{\textbf{Self-Generated Attacks Comparison.} Overview of the two types of our Self-Generated Attacks: \textbf{(a)} Class Based and \textbf{(b)} Reasoned Attacks. Class based attacks prompt a generative Large Vision Language Model (LVLM) (\eg LLaVA \cite{liu2023llava}) or text-image similarity based VL models (\eg CLIP \cite{radford2021learning})  about the most similar class to the ground truth and use that as an attack. Reasoned attacks prompt a generative LVLM to recommend an effective attack. The LVLM returns then a deceiving class and a reasoning.}
    \label{fig:figure_2}
\end{figure*}

Prior work on Typographic Attacks relies on using a single word as the deceiving text, failing to exploit the advanced language understanding and reasoning capabilities of Large Vision-Language Models (LVLMs). This simplistic approach limits the attack's potential to harm the model's predictive ability. For example, prior research has shown that incorporating a relevant sentence in response to an unsafe query can bypass LVLM safety mechanisms \cite{liu2023query}. Additionally, these attacks randomly sample a deceiving class from a predefined set of categories (Figure \ref{fig:figure_1} (a)), ignoring visual similarity, which could act as a stronger deceptive signal (\eg, a car brand more similar to Ford in Figure \ref{fig:figure_1} (a) than Rolls-Royce, a luxury brand, is likely a better choice).

To address these issues, We develop an experimental setup to test Large Vision Language Models susceptibility to typographic attacks that generalize to the recent Large Vision Language Models. The experimental setup uses five datasets and finds that Typographic attacks could reduce LVLM(s) performance by up to $60\%$. Moreover, as shown in Figure \ref{fig:figure_1} (b), we propose Self-Generated Attacks, a more effective class of attacks than Prior work Random Class attack \cite{azuma2023defense}. We identify two main methods to do so: 1) \textit{Class Based attacks} where the model (CLIP or LVLMs) is prompted to identify the most similar class to the ground truth and use that as a deceiving class (see Figure \ref{fig:figure_2} (a)) and 2) \textit{Reasoned attacks} where an advanced LVLM (\eg GPT4-V) is explicitly prompted to recommend the most confusing typographic attack against itself resulting in a deceiving class as well as an accompanying reasoning sentence to enhance the attack's credibility (see Figure \ref{fig:figure_2} (b)) Using our novel experimental setup, we show that our Self-Generated Attacks reduce performance by up to $60\%$, outperforming Random Class by $30\%$. Moreover, we show that Reasoned attacks that contain a motivating description outperform class-based attacks by $10\%$ on Large Vision Language Models whereas they are less effective against CLIP. This is evidence of the \textit{unique vulnerability of LVLM(s) to more verbose typographic attacks}.

Our contributions can be summarized: 

\begin{itemize}[leftmargin=*]
    \item We introduce a diverse and comprehensive experimental setup  that includes five datasets uniquely tailored to measure the susceptibility of the recently introduced Large Vision Language Models LVLM(s) as well as older VL models (\eg, CLIP) to Typographic Attacks. 
    \item We introduce a novel class of attacks: Self-Generated Typographic attacks, which use the model itself to generate the attack.
    \item We demonstrate how our Self-Generated attacks are the most effective attacks (can reduce performance by up to $60\%$). Moreover, reasoned attacks, which makes use of LVLM(s) language skills, are less effective on CLIP like models highlighting the unique vulnerability of LVLM(S). 
\end{itemize}

\section{Related Work}

\noindent\smallskip\textbf{Typographic Attacks.} Typographic attacks consist of misleading texts superimposed on images to distract vision and language models from their visual content. The susceptibility of vision and language models to Typographic attacks was noted in CLIP \cite{goh2021multimodal}. The authors found that by pasting the wrong class text on the image, they could increase misclassification errors on ImageNet \cite{5206848}. Since then, \cite{azuma2023defense} expanded the evaluation experimental setup s to datasets like OxfordPets \cite{parkhi2012cats}, StanfordCars \cite{krause20133d}, Flowers \cite{nilsback2008automated} and noted the same issue. We build on this work by developing a experimental setup  uniquely designed to test LVLM(s)  \cite{liu2023llava,zhu2023minigpt,dai2023instructblip,yang2023dawn} robustness to typographic attacks. Moreover, we introduce a novel and more effective typographic attack uniquely designed for LVLM(s).

\noindent\smallskip\textbf{Large Vision Language Models.}
Large Vision Language Models (LVLMs) \cite{liu2023llava,dai2305instructblip,zhu2023minigpt,alayrac2022flamingo} represent a new class of vision language models capable of impressive multimodal understanding skills. There are several ways to construct such models. For example, Flamingo \cite{alayrac2022flamingo} relies on a pretrained visual encoder and then trains a language model on interleaved visual and textual tokens. More recently, we have seen the introduction of models like LLaVA \cite{liu2023llava}, InstructBLIP \cite{dai2023instructblip}, and MiniGPT4 \cite{zhu2023minigpt} where a function is learned to project image representations from a visual encoder to the token space of a pretrained language model. Furthermore, these models were also trained to perform instruction following by either training on handpicked instruction templates \cite{dai2023instructblip}, or in works like \cite{liu2023llava,zhu2023minigpt} where captions in image-captions datasets are converted to instructions using conversational language models like GPT4 \cite{yang2023dawn}. In this work, we demonstrate how typographic attacks remain a significant concern for LVLM(s), just like their predecessor CLIP \cite{radford2021learning}.

\noindent\smallskip\textbf{Difference between Adversarial and Typographic Attacks.} Adversarial attacks \cite{yuan2021meta,rony2021augmented,zhu2021sparse,du2019query,chen2020universal} seek to develop an imperceptible noise, that disrupts the model visual recognition capabilities when added to the model. While our typographic attacks share a similar end goal to classic adversarial attacks, namely, disrupting model behavior, adversarial and typographic attacks are fundamentally different from each other.  Indeed, our attacks originate as a result of bias toward the textual signal in large foundation models, while adversarial attacks result from an artificial noise uniquely designed through the use of model intrinsic (\eg, gradients) to disrupt the fundamental mechanics of neural nets. Therefore, to develop such attacks, an attacker needs a sophisticated understanding of the intrinsic mechanics of neural nets such as \eg, Gradient  \cite{yuan2021meta,rony2021augmented,zhu2021sparse}, Meta Learning \cite{du2019query} or Attention maps \cite{chen2020universal} to name a few. However, the attacker, in our setup, only needs to know how to query the model.

\section{Problem Definition}

In this work, we are concerned with Typographic Attacks. Formally, assume we are given a dataset $D$ of image-target pairs $(X, Y)$. Now, given an image target pair $(x, y) \in D$, a Typographic attack algorithm $a$ \cite{azuma2023defense} takes in the pair and produces an image $\hat{x}$ with the text of deceiving class $\hat{y}$ pasted on it, \ie $a(x, y) = (\hat{x}, \hat{y})$ where $\hat{y} \neq y$. The attack is successful when the model uses the textual information of $\hat{y}$ from the attack to make its prediction rather than the visual content of $y$, hence mispredicting $y$ as $\hat{y}$.

\smallskip\noindent\textbf{Threat Model.} The threat model, we assume, involves a malicious but not necessarily sophisticated actor who leverages a vision language model to perform \textit{inference only} in order to obtain an attack. The model could be either a text-image similarity based vision language model (\eg, CLIP \cite{radford2021learning}) or a generative large vision language model (LVLM) (\eg, LLaVA \cite{liu2023llava}). This text is then superimposed on visual inputs intended for vision and language models. The simplicity and accessibility of this approach (especially when considering LVLM(s) which a user could interact with through language) make it a realistic and significant threat, as it requires minimal technical expertise or resources. In practice, this type of attack can be executed both digitally and physically. In a digital context, the adversary could edit images to include deceptive text before uploading them to social media platforms, online marketplaces, or any other digital repository where VL models are employed. Physically, the attack can be carried out by printing misleading textual descriptions and affixing them to real-world objects.

\begin{figure*}[t!]
    \centering
    \includegraphics[width=\linewidth]{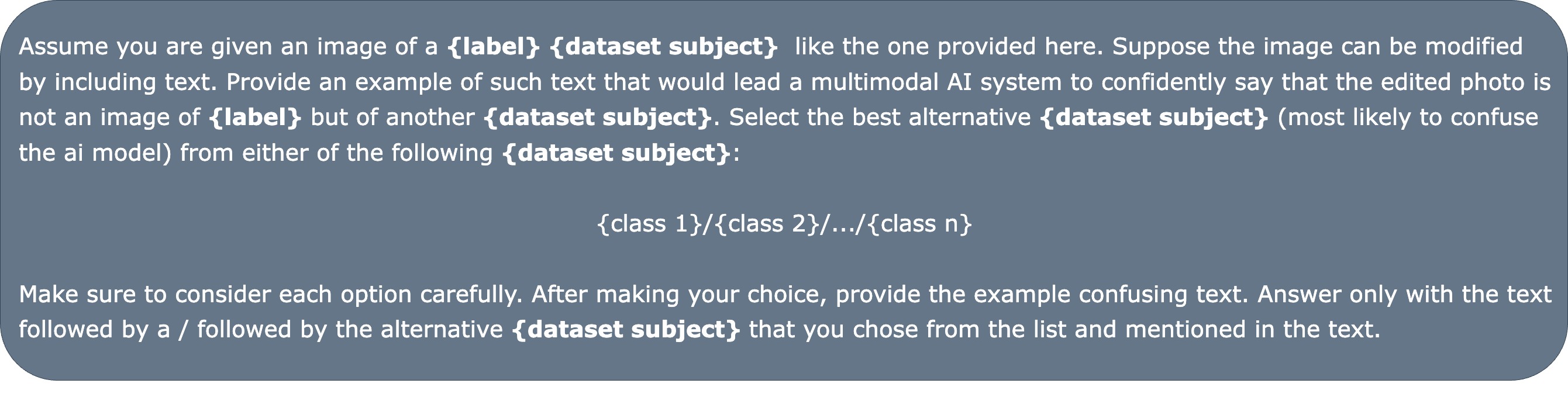}
    \caption{The prompt used to generate the Reasoned typographic attack with GPT-4V \cite{yang2023dawn}. \{dataset subject\} refers to the category of the dataset classes (\eg, Car Model for Cars). Refer to Section \ref{sec:self-generated} for further details.}
    \label{fig:figure_5}
\end{figure*}

\section{Self Generated Typographic Attacks}
\label{sec:self-generated}

Large Vision-Language Models (LVLMs) are capable of processing complex language, yet prior work on typographic attacks fails to exploit this as a potential vulnerability. These attacks are limited to using a single word, randomly chosen from a predefined set of categories, thereby missing the opportunity to leverage false reasoning to mislead the model \cite{azuma2023defense}.  Additionally, these attacks may not factor in visual similarity between the deceiving class and the true class, potentially reducing their effectiveness. For instance, randomly selecting \textit{hound} to mislabel a \textit{golden retriever} ignores the possibility of choosing a visually closer or contextually confusing breed, which could more effectively deceive the model. This oversight further limits the potential impact of prior typographic attacks.

Thus, we introduce a novel class of Typographic Attacks: Self-Generated Attacks, which use vision-language model itself to create the attack. Then, we identify two main attacks under this class: \textit{Class Based Attacks} (Section \ref{sec:class_based_attacks}) and \textit{Reasoned Attacks} (Section \ref{sec:desc_based_attacks}).

\subsection{Self Generated Class-Based Attacks}
\label{sec:class_based_attacks}

Class-Based Attacks are based on a straightforward observation: a visually similar deceiving class to the target class is likely more effective at misleading the model than a random class. Leveraging this insight, we propose Class-Based Attacks where we utilize the a vision language model to identify which deceiving class is most visually similar to the ground truth.  Note how this method can apply to both image-text similarity based VL models (\eg CLIP \cite{radford2021learning}) as well as generative large vision language models (\eg LLaVA \cite{liu2023llava}) as shown in Figure \ref{fig:figure_2} (a). For models like CLIP, we encode each image through CLIP's visual encoder, and then obtain the class that has the highest similarity score to the image that is not the ground truth class. For the LVLM, we simply prompt the model following  Figure \ref{fig:figure_2} (a)  to determine the closest deceptive class based on visual attributes. By prompting the model itself to identify a visually similar deceiving class, we harness its ability to recognize and compare fine-grained visual details. This approach enhances the likelihood of a successful misclassification, as the model is more prone to confuse the target class with a visually similar one.

\subsection{Self Generated Reasoned Attacks}
\label{sec:desc_based_attacks}

While Class-Based Attacks (Section \ref{sec:class_based_attacks}) rely on a stronger prior, leveraging visual similarity to mislead the model, they fail to exploit the sophisticated language capabilities inherent in LVLMs. These models can process and incorporate rich, Reasoned language beyond simple class labels when making predictions. Recognizing this, we hypothesized that a Reasoned attack, which provides a rationale for the misleading class, could be more effective than merely pasting a deceptive class label. To test this hypothesis, we explored the potential of Reasoned attacks, where the misleading text not only specifies a different class but also includes a reasoning that supports the deception. Our approach involves asking the LVLM itself to generate these attacks. By prompting the model to recommend an attack strategy against its own predictions, as demonstrated in Figure \ref{fig:figure_2} (b), we obtain both a misleading class and a detailed, persuasive description designed to confuse the model. This self-generated attack method taps into the model's advanced language understanding, creating more nuanced and effective typographic attacks. By doing so, we aim to evaluate whether these richer, more contextually grounded attacks are superior in deceiving LVLMs compared to simpler class-based manipulations (Section \ref{sec:class_based_attacks}). Examine Figure \ref{fig:figure_5} for a template of the prompt used to obtain these attacks. We find it important to instruct the model to ``consider each option carefully'' to obtain the most effective attacks.

\section{Typographic Attacks against Large Vision Language Models}
\label{sec:def}

Prior work on Typographic Attacks has been limited to the vision language model CLIP \cite{azuma2023defense}. However, recently, the field has made rapid progress on a new class of vision-language models: Large Vision Language Models \cite{liu2023llava,zhu2023minigpt,dai2023instructblip,yang2023dawn} which rely on a Large Language Model. This property enables an impressive conversational ability which expands the pool of potential model users. However, typographic attacks against these models remain understudied. Therefore, in this work, seek to address this problem by developing an experimental setup to test the susceptibility vision language models to typographic attacks that generalize to LVLMs.  Therefore, inspired by recent work on evaluating LVLM(s) \cite{fu2023mme,xu2023lvlm}, we propose the following setup.

Assume we are given a set of typographic attack algorithms $a \in A$, each takes in a pair of image-target, \ie $(x, y)$, and produces the deceiving target-image pair $(\hat{x}, \hat{y})$. To compare the impact of the typographic attacks on LVLM(s), we confront the LVLM with each algorithm set of manipulated images and ask it to choose the correct answer $y$ among the set of choices $C$ comprised of the deceiving classes produced by each algorithm as well as the ground truth $y$, \ie $C = \{y\} \cup \{a(x, y)\} \quad \forall a \in A$. For example, if the image is labeled with $Jeep$ and two algorithms result in attacks that contain $Audi$ and $Fiat$, then we instruct the model to choose between $\{Jeep, Audi, Fiat\}$. We shuffle the ordering of each option in the question prompt to avoid model bias to any answer order. 

More formally, assuming $C$ is the set of classes that the typographic attack algorithm produces, and $y$ is the image ground truth, then we pose the LVLM with the following question: 

\begin{quote}
    Select the correct \{Dataset Subject\} pictured in the image: (1) \{$C[0]$\}, (2) \{$C[1]$\}, (3) $y$ ... (N) \{Class $N$\}. Answer with either (1) or (2) ... (N) only.
\end{quote}

We then test whether the model answer contains the correct choice (in this case (3) ). Moreover, we ensure that the order of $y$ is randomized in the answer options to avoid model bias to particular answer numbers. Note that the setup could be easily used to test image-text similarity based model (\eg CLIP \cite{radford2021learning}) by simply encoding each option through the text encoder and then retrieving the option with highest similarity score to the image.

\begin{table*}[th!]
\centering

\resizebox{\textwidth}{!}{\begin{tabular}{l|c|c|ccc|cc}
\hline
 & \multicolumn{1}{l|}{No Typographic} & \multicolumn{1}{l|}{Random Class  } & \multicolumn{3}{c|}{Class Based Attacks (ours)} & \multicolumn{2}{c}{Reasoned Attacks (ours)} \\ 
 & Attack & Attack$\downarrow$  & {VE} $\downarrow$ & {LLM} $\downarrow$ & {LVLM} $\downarrow$ & {LLM} $\downarrow$ & {LVLM} $\downarrow$ \\ \hline
CLIP & $63.3$ & $45.5$ &  \boldmath  $12.0$ & $42.2$ & $32.9$ & $26.6$ & $20.4$  \\ \hline 
GPT4-V   & $72.7$ & $66.0$ & $38.9$ & $57.8$ & $50.9$ & $58.1$ &  \boldmath  $31.8$  \\ 
LLaVA 1.5   & $50.8$ & $27.3$ & $18.3$ & $18.2$ & $13.2$ & $11.5$ &  \boldmath  $9.9$  \\ 
InstructBlip   & $60.2$ & $26.8$ & $20.6$ & $23.0$ & $22.2$ &  \boldmath  $13.9$ & $14.9$  \\ 
MiniGPT4-2    & $27.7$ & $25.6$ & $25.7$ & $24.6$ & $25.3$ & $23.7$ & 
  \boldmath  $22.4$  \\ 
Avg Accuracy & $54.9$ & $38.2$ & $23.1$ & $33.2$ & $28.9$ & $26.8$ &  \boldmath  $19.9$  \\ \cdashline{1-8} 
Avg \% Accuracy Drop $\uparrow$ & $-$ & $29.3$ & $52.9$ & $38.2$ & $44.8$ & $49.3$ & \boldmath $59.8$  \\ \hline 
\end{tabular}}

\caption{Comparison between the effect of different typographic attacks (Random Class \cite{azuma2023defense} and our Self-Generated Attacks) on the accuracy of the following Large Vision Lanugage Models: GPT-4V \cite{yang2023dawn}, LLaVA 1.5 \cite{liu2023llava}, MiniGPT4-2 \cite{zhu2023minigpt}, and  InstructBLIP \cite{dai2023instructblip} as well as CLIP \cite{radford2021learning}. In addition, we report Avg Accuracy and Avg \% Accuracy Drop as defined in Section \ref{sec:exp}.   Refer to Section \ref{sec:exp} for further discussion.}
\label{tb:table_1_main_results}
\end{table*}

Finally, \cite{azuma2023defense} paste the attack at a random location on the image. However, this might occlude important visual cues (\eg, a car logo when predicting the car model). To avoid this issue, we add a white space at the bottom and top of the image to allow for textual attacks. Refer to the Appendix for further details.

\section{Experiments}
\label{sec:exp}

\noindent\textbf{Typographic Attacks Algorithms.}  We compare the effect of  \textit{Random Class Attack} \cite{azuma2023defense}  where a randomly sampled class is pasted on the image to our \textit{Class Based Attack} and \textit{Reasoned Attacks}  (Section \ref{sec:self-generated}). For Class Based Attacks, we use  the Visual Encoder (\textit{VE}) underlying LLaVA 1.5 (CLIP) to retrieve the most similar class, the Large Language Model \textit{LLM} underlying LLaVA 1.5 \cite{liu2023llava}  which does not factory visual info, as well as LLaVA itself as outlined in Section \ref{sec:class_based_attacks}. For Reasoned Attacks, we note that open source models fail to effectively respond to our attack query. Therefore, we use GPT-4V \cite{yang2023dawn}, a more capable model. We query both the LLM underlying GPT4-V which does not factor visual info as well as GPT4-V itself. 

\noindent\textbf{Datasets.} We provide detailed reasonings of the datasets used in our experiments, covering a diverse array of domains. The OxfordPets dataset \cite{parkhi2012cats} consisting of 37 pet breeds of various cat and dog breeds. The StanfordCars dataset \cite{krause20133d} includes 196 car models. The Flowers dataset \cite{nilsback2008automated} contains 102 flower species. The Aircraft dataset \cite{maji2013fine} features 100 aircraft models. Finally, the Food101 dataset \cite{bossard2014food} comprises  101 food dishes covering a variety of cuisines and presentation styles. Moreover, note that we limit the number of samples in the test set per dataset to 1000 samples. This is due to the computational and monetary costs associated with evaluating GPT-4V.

\noindent\textbf{Models.} We test on four of the recent large vision language models (LVLMs). Namely, we test 1) LLaVA 1.5 \cite{liu2023llava}: which combines the visual encoder CLIP \cite{radford2021learning} with the Language Model Vicuna \cite{peng2023instruction} and train the final model on a multimodal language-image data where the language follows an instruction format created using GPT-4 \cite{achiam2023gpt} 2) MiniGPT4 \cite{zhu2023minigpt} which follows a similar structure to LLaVA. However, the visual encoder is ViT backbone \cite{fang2023eva} and the instruction data used to fine tune the model is based on manually defined templates. 3) InstructBLIP \cite{dai2023instructblip} which used the pretrained vision-language model BLIP-2 \cite{li2023blip} and fine tune to follow instruction on a multimodal instruction dataset based on manually crafted instructions  and 4)  GPT-4V \cite{yang2023dawn} a close source vision-language model capable of instruction following by OpenAI. Finally, even though our experimental setup  is tailored to test LVLM(s) performance, we provide the performance of CLIP \cite{radford2021learning} to be complete. 

\noindent\textbf{Metrics.} We use accuracy; the number of times each model chooses the correct answer from the prompt outlined in Section \ref{sec:def}. Moreover, we report Average \textit{\% Accuracy Drop} which we define as: 

\begin{equation*}
    \text{\% Accuracy Drop} = \frac{Acc_{Baseline} - Acc_{a}}{Acc_{Baseline}}
\end{equation*}

for a given typographic Attack algorithm $a$ where a lower number indicates a more effective attack.

\begin{figure*}[th!]
    \centering
    \includegraphics[width=0.85\linewidth]{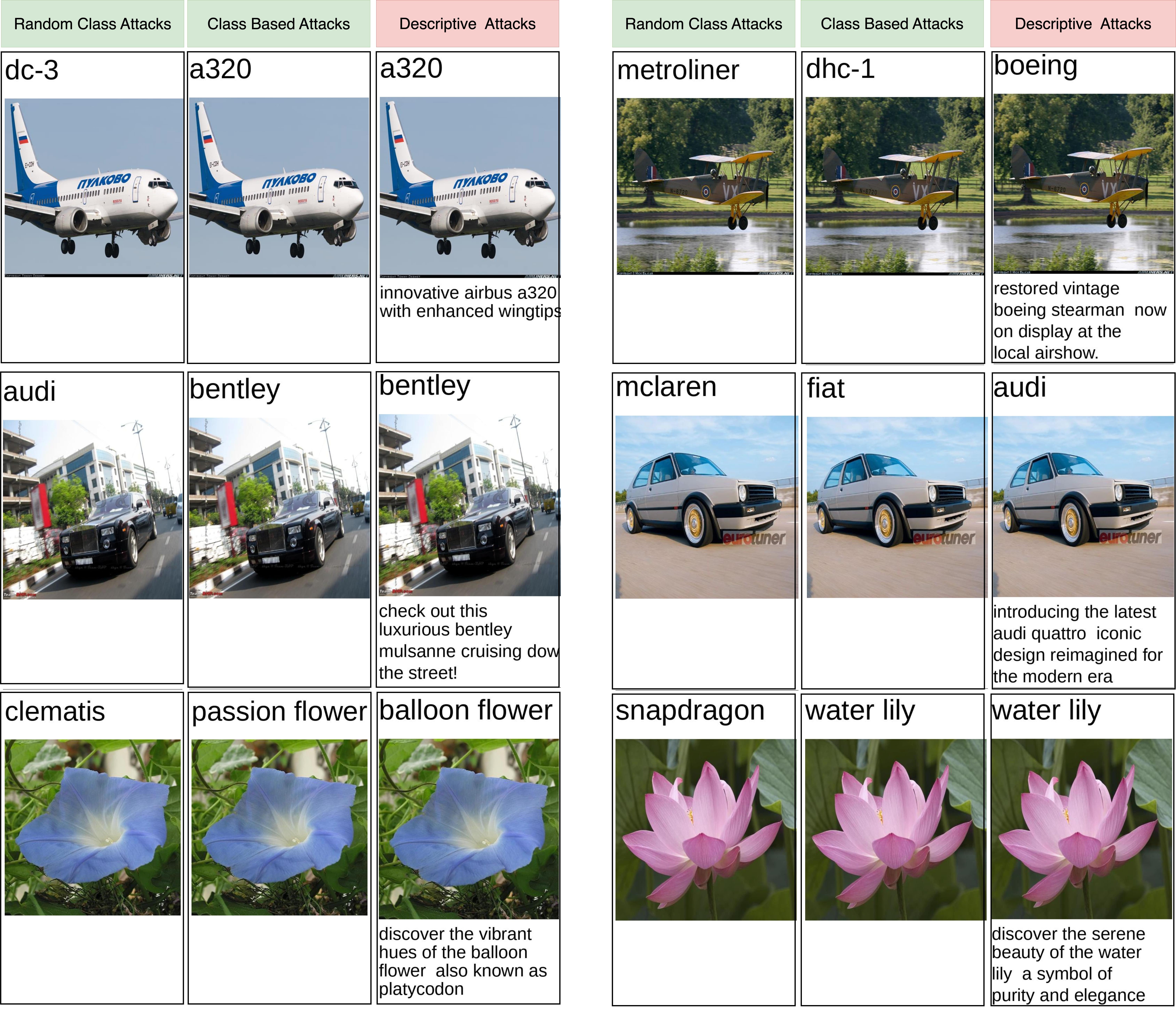}
    \caption{Qualitative examples where our Reasoned attacks (Column 3) cause the model to misclassify the mage while our Class Based Attacks (Column 2) and Random Class Attacks \cite{azuma2023defense} fail to do so on the five datasets used in our experimental setup, namely:  The Aircraft dataset \cite{maji2013fine} (Row 1), The StanfordCars dataset \cite{krause20133d} (Row 2), The Flowers dataset \cite{nilsback2008automated} (Row 3), The OxfordPets dataset \cite{parkhi2012cats} (Row 4), and  Food101 dataset \cite{bossard2014food}  (Row 5). Refer to Section \ref{sec-apdx:qual_examples} for Discussion. }
    \label{fig:figure_6}
    \vspace{-2mm}
\end{figure*}

\subsection{Results}
\label{sec:main_results}

Table \ref{tb:table_1_main_results} reports the effect of different Typographic Attacks averaged over all datasets. We find Reasoned Attacks using the LVLM GPT-4V outperform Random Class attacks \cite{azuma2023defense} with respect to Avg \% Accuracy Drop by $30\%$ as well as the best method under Class Based Attacks (VE) by $7\%$. This is likely because, unlike Random Class and Class Based Attacks where the attack consists of a class only, Reasoned  Attacks also paste a motivating reasoning. Nevertheless, note how Reasoned attacks are less effective on CLIP; the class based attack using CLIP itself (VE) is the most effective. This is likely because CLIP doesn't possess the advanced language understanding skills like LVLM(s) that could incorporate the reasoning. Moreover, note how for LLaVA and InstructBLIP, Random Class is a fairly effective attack; the models lose almost half of their base performance. However, GPT4-V only loses about $9\%$ of its performance with Random Class but loses 6x times that with our Reasoned attack. This is likely because GPT4-V has better reasoning capabilities \cite{fu2023mme} and, thus, is less likely to be deceived by a random class. 

Despite having the lowest classification performance with no typographic attack, MiniGPT-4 is most robust against Typographic Attacks; it only loses about $20\%$ of its performance with the most effective attack. This is likely because MiniGPT4 shows poorer performance in optical character recognition (OCR) \cite{liu2023hidden}, which makes it less likely to incorporate the attack into its prediction. GPT-4V comes second, which loses $56\%$ of its performance with the most effective attack, which is significantly lower (and hence more robust) than either LLaVA or InstructBLIP, which lose $75-80\%$ of their performance with the most effective attack.

\subsection{Qualitative Examples}
\label{sec-apdx:qual_examples}
In this Section, we provide qualitative examples where our Self Generated Reasoned Attacks outperform both our Self Generated Class Attack and Random Class Attacks \cite{azuma2023defense}. Refer to Figure \ref{fig:figure_6} for results. Note how our method (Self Generated Attacks) effectively generalizes to datasets of different domains. Indeed, Class-Based Attacks are able to effectively recommend similar and hence more effective classes (Audi, which is more similar to Volkswagen than Mcalren (Column2)), and Reasoned attacks are able to recommend a convincing deceiving reasoning as well as a deceiving class across domains. For example, the Reasoned Attack in Column (6) and Row (2) justifies the old look of the 
Audi with ``re-imagining for the modern era'' and hence fooling the model while other attacks do not manage to.

\subsection{Ablations}
\label{sec:ablations_results}

\begin{figure}
    \centering
    \includegraphics[width=\linewidth]{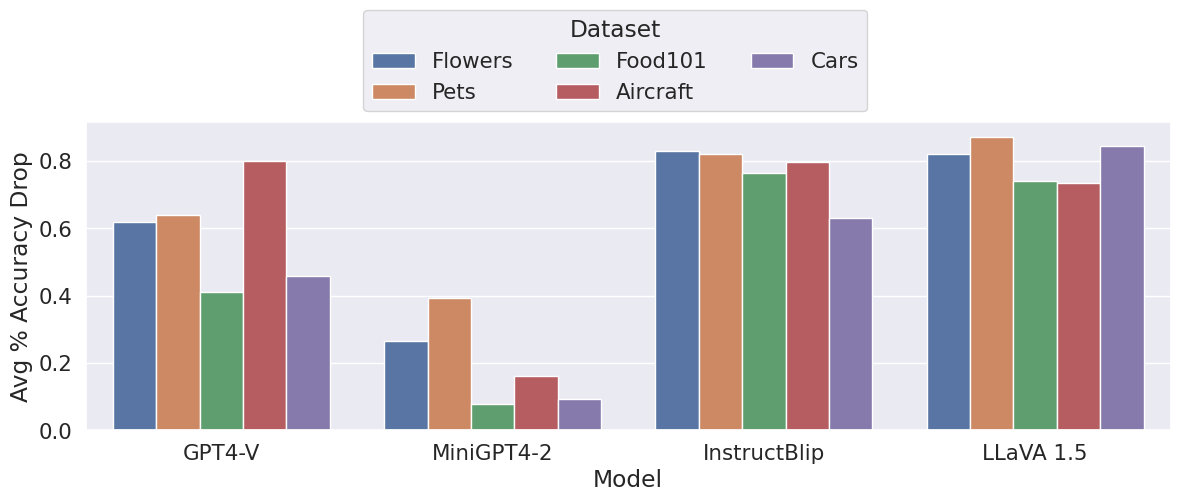}
    \caption{Comparing the Avg \% Accuracy drop of our Reasoned Attack on  1) GPT4-V\cite{yang2023dawn} 2) MiniGPT-4 \cite{zhu2023minigpt}  3) InstructBLIP \cite{dai2023instructblip} and 4) LLaVA1.5 \cite{liu2023llava} on the five datasets that comprise our experimental setup  1) Flowers  2) Food101   3) Cars  4) Pets  and 5) Aircraft. Refer to Section \ref{sec:ablations_results} for further discussion.  }
    \label{fig:figure_8}
    \vspace{-3mm}
\end{figure}

\smallskip\noindent\textbf{Breaking down Performance by Dataset.}  Our experimental setup  is comprised of five different datasets, each constituting a unique visual domain, namely  1) Flowers, 2) Food101, 3) Cars, 4) Pets, and 5) Aircraft. Thus, in this experiment, we seek to understand if certain domains are more susceptible to typographic than others. Figure \ref{fig:figure_8} compares the effect of our Self-Generated Reasoned attack on the five datasets that comprise our experimental setup . Overall, note that the Self-Generated Reasoned attack is consistently effective against the Aircraft domain across models. This is likely because the fine grained differences between aircrafts are not well represented in the pretraining datasets of models compared to the other domains (\eg, flowers or pets). On the other hand, attacks against the food domain are less effective overall than those against other domains across models. This is likely because food images are more represented in pretraining data. More interestingly, the distribution of the attack effects changes substantially per domain across models. For example, while attacks against Cars are more effective than other domains on LLaVA and InstructBLIP, they are less effective on GPT4-V. This likely means that GPT4-V is more capable of correctly identifying the correct car category, and thus, the deceptive textual signal is less effective. 

\begin{figure}[t!]
    \centering
    \includegraphics[width=\linewidth]{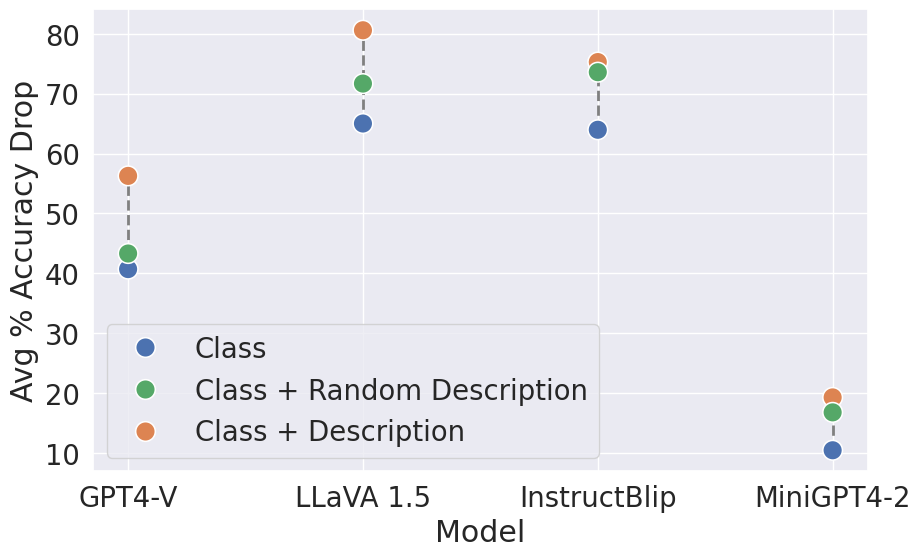}
    \caption{Comparing the effect of descriptions produced by the Recommended Attacks on performance. Refer to Section \ref{sec:ablations_results} for further discussion.}
    \label{fig:figure_3}
    \vspace{-4mm}
\end{figure}

\begin{figure}[t!]
    \centering
    \includegraphics[width=\linewidth]{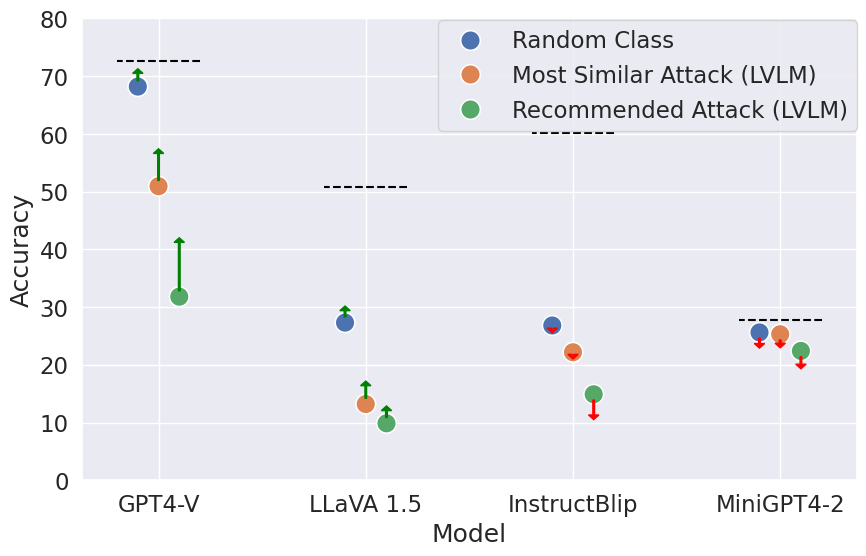}
    \caption{Comparing the effect of prompting the LVLM to ignore the typographic attack. Green arrows refer to gains in performance, and red refer to drop. Black dashed line refers to baseline performance with no attacks. Refer to Section \ref{sec:ablations_results} for further discussion.}
    \label{fig:figure_4}
\end{figure}

\smallskip\noindent\textbf{Ablating the reasoning in Reasoned Attacks.} We test the effect of reasonings in our Reasoned  Attacks (Section \ref{sec:self-generated}) on the overall performance. Figure \ref{fig:figure_3} reports the Avg \% Accuracy Drop with Reasoned Attacks when using no reasoning, a random reasoning from another image, and the reasoning for that image. Note that using the image reasoning is the most effective at reducing performance across models. Most notably, we find a random reasoning attack on GPT-4 performs almost the same as having no reasoning. This is likely because of the strong reasoning capabilities of GPT4-V which likely enables it to realize the discrepancy between the reasoning and the content of the image. We find a different trend with the other models where a random reasoning is more powerful than the class by itself. This is likely due to these models weaker reasoning capabilities. Nevertheless, the correct reasoning attack (Blue) maintains the strongest drop in performance across models.

\smallskip\noindent\textbf{Can LVLM(s) Ignore the Attack?} Could Typographic Attacks be mitigated by simply prompting the LVLM to ignore the text in the image? Note results in Figure \ref{fig:figure_4}. Models fail to gain back their base performance (dashed lines) without a Typographic Attack. Nevertheless, some models still make some gains. Indeed, we see the most improvements from  GPT4-V while we see mild improvements from LLaVA. More surprisingly, we see a decline in performance from InstructBLIP and MiniGPT4-2, which indicates that these models are not capable of executing this instruction.  Overall, the results indicate that future work should be pay a greater attention to  mitigating Typographic Attacks in LVLM(s).

\section{Conclusion}

In this paper, we introduced an experimental setup for typographic attacks that generalizes to large vision and language models (LVLMs), systematically exposing their vulnerabilities and showing how such attacks can severely degrade performance. We proposed a novel class of Self-Generated Typographic Attacks, which exploit LVLMs' advanced language reasoning capabilities to generate deceptive text. Unlike prior work, these attacks leverage the models' own reasoning, making them significantly more effective, particularly on LVLMs with richer language skills—capabilities absent in models like CLIP, which lack sophisticated reasoning. Our experiments revealed that Self-Generated Typographic Attacks outperform prior approaches, especially when incorporating reasoned language constructs. Extensive ablations highlighted: 1) the importance of descriptions in Reasoned Attacks, 2) model vulnerability even when prompted to ignore text, and 3) variations in attack effectiveness across domains. These results underscore the critical need for defenses against attacks that exploit LVLMs' language reasoning abilities.

\smallskip\noindent\textbf{Limitations: } In this work, we study Typographic Attacks against Large Vision Language Models (LVLM(s)). While we consider four of the latest and most popular models, new open-source models are constantly being introduced. Nevertheless, by shedding light on Typographic Attacks on these four models, we hope that this prompts the community to pay greater attention to this problem when developing and evaluating new models. Moreover, in our work, we focus on deceiving the model about the type of various visual entities in an image (\eg, car model, flower breed). However, LVLM(s) are versatile and can be used to understand more aspects of the visual entities such as their  color, action, position, etc.  Thus, Future work could benefit from extending our study to such cases.

\bibliography{custom}

\appendix

\begin{figure}[h!]
    \centering
    \includegraphics[width=0.9\linewidth]{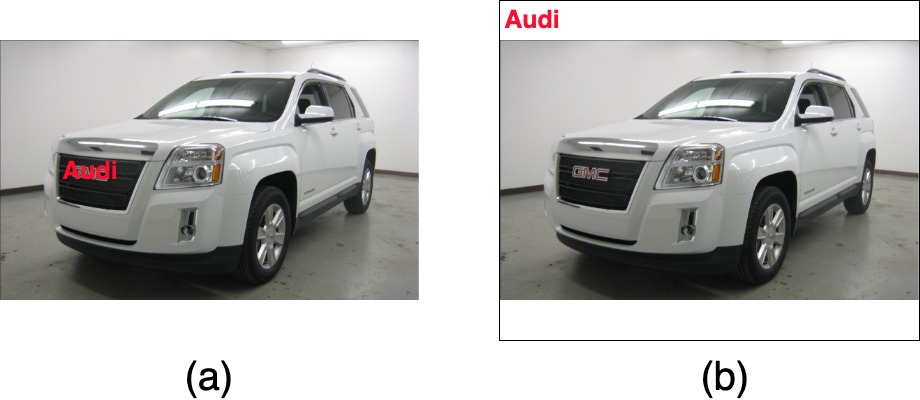}
    \caption{Comparing prior work process for testing typographic attacks on an image (a) and ours (b) where add white space at the bottom and top of the image to allow for the attacks. Our format avoids the scenario where the typographic attack hides an important visual cue (care model) required to make the prediction. Refer to Section \ref{sec:def} for discussion. }
    \label{fig:figure_7}
\end{figure}

\begin{table*}[th!]
\centering

\begin{subtable}{\textwidth}
\centering

\resizebox{\textwidth}{!}{\begin{tabular}{l|c|c|ccc|cc}
\hline
 & \multicolumn{1}{l|}{No Typographic} & \multicolumn{1}{l|}{Random Class  } & \multicolumn{3}{c|}{Class Based Attacks (ours)} & \multicolumn{2}{c}{Reasoned Attacks (ours)} \\ 
 & Attack & Attack \cite{azuma2023defense} $\downarrow$  & {VE} $\downarrow$ & {LLM} $\downarrow$ & {LVLM} $\downarrow$ & {LLM} $\downarrow$ & {LVLM} $\downarrow$ \\ \hline
CLIP & $25.4$ & $18.3 $ & $6.4 $ & $20.1 $ & $14.9 $ & $12.8 $ & $9.8 $  \\ \hline 
GPT4-V & $44.6$ & $37.4 $ & $15.8 $ & $28.4 $ & $21.8 $ & $30.6 $ & $8.9 $  \\ 
LLaVA 1.5 & $26.1$ & $11.8 $ & $7.2 $ & $7.3 $ & $5.3 $ & $4.2 $ & $6.9 $  \\ 
InstructBlip & $26.1$ & $4.7 $ & $6.2 $ & $3.3 $ & $4.0 $ & $3.8 $ & $5.3 $  \\ 
MiniGPT4-2 & $19.7$ & $19.7$ & $20.0 $ & $21.2 $ & $19.3 $ & $18.4 $ & $16.5 $  \\ 
Average Accuracy & $28.4$ & $18.4$ & $11.1$ & $16.0$ & $13.0$ & $13.9$ & $9.5$  \\ \cdashline{1-8} 
Average Accuracy Drop $\uparrow$ & $-$ & $36.1$ & $57.3$ & $41.8$ & $51.8$ & $51.4$ & $62.2$  \\ \hline 
\end{tabular}}

\caption{Aircraft \cite{maji2013fine} }
\end{subtable}

\begin{subtable}{\textwidth}
\centering

\resizebox{\textwidth}{!}{\begin{tabular}{l|c|c|ccc|cc}
\hline
 & \multicolumn{1}{l|}{No Typographic} & \multicolumn{1}{l|}{Random Class  } & \multicolumn{3}{c|}{Class Based Attacks (ours)} & \multicolumn{2}{c}{Reasoned Attacks (ours)} \\ 
 & Attack & Attack \cite{azuma2023defense} $\downarrow$  & {VE} $\downarrow$ & {LLM} $\downarrow$ & {LVLM} $\downarrow$ & {LLM} $\downarrow$ & {LVLM} $\downarrow$ \\ \hline
CLIP & $62.7$ & $47.2 $ & $13.7 $ & $55.4 $ & $34.4 $ & $33.2 $ & $24.1 $  \\ \hline 
GPT4-V & $81.0$ & $74.1 $ & $58.3 $ & $70.4 $ & $62.7 $ & $70.4 $ & $43.9 $  \\ 
LLaVA 1.5 & $69.0$ & $43.0 $ & $31.4 $ & $27.8 $ & $21.4 $ & $10.8 $ & $10.8 $  \\ 
InstructBlip & $85.0$ & $53.1 $ & $47.6 $ & $42.6 $ & $45.0 $ & $29.7 $ & $31.3 $  \\ 
MiniGPT4-2 & $33.4$ & $33.4$ & $33.5 $ & $29.2 $ & $34.3 $ & $34.0 $ & $30.3 $  \\ 
Average Accuracy & $66.2$ & $50.1$ & $36.9$ & $45.1$ & $39.5$ & $35.6$ & $28.1$  \\ \cdashline{1-8} 
Average Accuracy Drop $\uparrow$ & $-$ & $21.7$ & $40.9$ & $29.4$ & $36.3$ & $41.6$ & $52.9$  \\ \hline 
\end{tabular}}
\caption{StanfordCars \cite{krause20133d}}

\end{subtable}

\begin{subtable}{\textwidth}
\centering

\resizebox{\textwidth}{!}{\begin{tabular}{l|c|c|ccc|cc}
\hline
 & \multicolumn{1}{l|}{No Typographic} & \multicolumn{1}{l|}{Random Class  } & \multicolumn{3}{c|}{Class Based Attacks (ours)} & \multicolumn{2}{c}{Reasoned Attacks (ours)} \\ 
 & Attack & Attack \cite{azuma2023defense} $\downarrow$  & {VE} $\downarrow$ & {LLM} $\downarrow$ & {LVLM} $\downarrow$ & {LLM} $\downarrow$ & {LVLM} $\downarrow$ \\ \hline
CLIP & $68.7$ & $36.9 $ & $4.3 $ & $33.9 $ & $29.5 $ & $27.2 $ & $16.6 $  \\ \hline 
GPT4-V & $74.5$ & $65.5 $ & $33.4 $ & $56.8 $ & $50.2 $ & $66.3 $ & $28.3 $  \\ 
LLaVA 1.5 & $38.3$ & $10.8 $ & $9.3 $ & $5.3 $ & $6.5 $ & $6.5 $ & $6.8 $  \\ 
InstructBlip & $48.3$ & $15.5 $ & $11.6 $ & $17.9 $ & $16.4 $ & $5.9 $ & $8.2 $  \\ 
MiniGPT4-2 & $20.1$ & $17.8 $ & $18.5 $ & $17.5 $ & $17.9 $ & $16.3 $ & $14.7 $  \\ 
Average Accuracy & $50.0$ & $29.3$ & $15.4$ & $26.3$ & $24.1$ & $24.4$ & $14.9$  \\ \cdashline{1-8} 
Average Accuracy Drop $\uparrow$ & $-$ & $41.9$ & $61.7$ & $47.3$ & $49.9$ & $52.2$ & $65.9$  \\ \hline 
\end{tabular}}
\caption{Flowers \cite{nilsback2008automated}, }

\end{subtable}

\begin{subtable}{\textwidth}
\centering

\resizebox{\textwidth}{!}{\begin{tabular}{l|c|c|ccc|cc}
\hline
 & \multicolumn{1}{l|}{No Typographic} & \multicolumn{1}{l|}{Random Class  } & \multicolumn{3}{c|}{Class Based Attacks (ours)} & \multicolumn{2}{c}{Reasoned Attacks (ours)} \\ 
 & Attack & Attack \cite{azuma2023defense} $\downarrow$  & {VE} $\downarrow$ & {LLM} $\downarrow$ & {LVLM} $\downarrow$ & {LLM} $\downarrow$ & {LVLM} $\downarrow$ \\ \hline
CLIP & $78.9$ & $66.1 $ & $24.4 $ & $56.0 $ & $46.5 $ & $41.6 $ & $30.6 $  \\ \hline 
GPT4-V & $82.9$ & $77.6 $ & $45.5 $ & $68.1 $ & $58.6 $ & $75.1 $ & $48.8 $  \\ 
LLaVA 1.5 & $71.4$ & $54.4 $ & $32.6 $ & $41.3 $ & $26.2 $ & $30.6 $ & $18.5 $  \\ 
InstructBlip & $76.0$ & $38.2 $ & $21.3 $ & $33.0 $ & $24.7 $ & $17.8 $ & $17.9 $  \\ 
MiniGPT4-2 & $33.7$ & $32.5 $ & $32.9 $ & $32.8 $ & $32.4 $ & $29.2 $ & $31.1 $  \\ 
Average Accuracy & $68.6$ & $53.8$ & $31.3$ & $46.2$ & $37.7$ & $38.8$ & $29.4$  \\ \cdashline{1-8} 
Average Accuracy Drop $\uparrow$ & $-$ & $19.9$ & $48.6$ & $29.7$ & $41.0$ & $40.8$ & $52.2$  \\ \hline

\end{tabular}}
\caption{Food101 \cite{bossard2014food}. }

\end{subtable}

\begin{subtable}{\textwidth}
\centering

\resizebox{\textwidth}{!}{\begin{tabular}{l|c|c|ccc|cc}
\hline
 & \multicolumn{1}{l|}{No Typographic} & \multicolumn{1}{l|}{Random Class  } & \multicolumn{3}{c|}{Class Based Attacks (ours)} & \multicolumn{2}{c}{Reasoned Attacks (ours)} \\ 
 & Attack & Attack \cite{azuma2023defense} $\downarrow$  & {VE} $\downarrow$ & {LLM} $\downarrow$ & {LVLM} $\downarrow$ & {LLM} $\downarrow$ & {LVLM} $\downarrow$ \\ \hline
CLIP & $80.6$ & $59.0 $ & $11.4 $ & $45.7 $ & $39.1 $ & $18.2 $ & $21.3 $  \\ \hline 
GPT4-V & $80.5$ & $75.4 $ & $41.8 $ & $65.2 $ & $61.3 $ & $48.4 $ & $29.1 $  \\ 
LLaVA 1.5 & $49.2$ & $16.5 $ & $11.2 $ & $9.6 $ & $6.6 $ & $5.4 $ & $6.3 $  \\ 
InstructBlip & $65.6$ & $22.5 $ & $16.3 $ & $18.4 $ & $20.8 $ & $12.2 $ & $11.8 $  \\ 
MiniGPT4-2 & $31.9$ & $24.7 $ & $23.8 $ & $22.6 $ & $22.6 $ & $20.9 $ & $19.4 $  \\ 
Average Accuracy & $61.6$ & $39.6$ & $20.9$ & $32.3$ & $30.1$ & $21.0$ & $17.6$  \\ \cdashline{1-8} 
Average Accuracy Drop $\uparrow$ & $-$ & $37.6$ & $62.4$ & $48.8$ & $51.9$ & $64.5$ & $69.2$  \\ \hline 
\end{tabular}}
\caption{OxfordPets \cite{parkhi2012cats}, }

\end{subtable}

\caption{Comparison between the effect different typographic attacks (Random Class \cite{azuma2023defense} and our Self-Generated Attacks) on Large Vision Lanugage Models: GPT-4V \cite{yang2023dawn}, LLaVA 1.5 \cite{liu2023llava}, MiniGPT4-2 \cite{zhu2023minigpt}, and  InstructBLIP \cite{dai2023instructblip} across various Datasets. Refer to Section \ref{sec-apdx:dataset_breakdown} for further discussion.}
\label{tb:results_per_dataset}

\end{table*}

\section{Results for each dataset}
\label{sec-apdx:dataset_breakdown}
In Section 6, we reported the results of each typographic attack averaged over all the datasets in our experimental setup. In this Section, we break down the results per dataset. Refer to Table \ref{tb:results_per_dataset} for results. Note that as we discuss in Section 6, our  Self Generated typographic attacks (Section 4), including class-based and Reasoned attacks, are consistently more effective than prior work random class attacks, in reducing model performance. Moreover, on average, Reasoned attacks are more effective than class-based attacks at reducing model performance. This is likely due to their use of deceiving prompts, which harnesses the sophisticated language understanding capabilities of LLM(s).

\section{Attack Location}

Prior work \cite{azuma2023defense} paste the attack at a random location on the image. However, this might occlude important visual cues (\eg, a car logo when predicting the car model). To avoid this issue, we add a white space at the bottom and top of the image to allow for textual attacks. Examine Figure \ref{fig:figure_7} for an illustrative example. Note how prior work attack (a) occludes the car logo, which is important to make the prediction about the car model. However, our method  Figure \ref{fig:figure_7} (b) avoids this by allocating white space at the bottom and top of the image for the textual attack. 

\end{document}